\documentclass{article}

\usepackage{microtype}
\usepackage{graphicx}
\usepackage{subfigure}
\usepackage{booktabs} % for professional tables

\usepackage{hyperref}

\newcommand{\specialcell}[2][c]{%
  \begin{tabular}[#1]{@{}c@{}}#2\end{tabular}}
\usepackage[accepted]{icml2024}

\usepackage{amsmath}
\usepackage{amssymb}
\usepackage{mathtools}
\usepackage{amsthm}
\usepackage{multirow}
\usepackage[capitalize,noabbrev]{cleveref}

\theoremstyle{plain}

\theoremstyle{definition}

\theoremstyle{remark}

\usepackage[textsize=tiny]{todonotes}
\usepackage{graphicx}	% Including figure files
\usepackage{amsmath}	% Advanced maths commands
\usepackage{bm}
\usepackage{float}

\icmltitlerunning{Classifier-Based Anomaly Detection}

\begin{document}

\twocolumn[
\icmltitle{A Classifier-Based Approach to Multi-Class Anomaly Detection Applied to Astronomical Time-Series}



\graphicspath{ {./images/} }

\icmlsetsymbol{equal}{*}

\begin{icmlauthorlist}
\icmlauthor{Rithwik Gupta}{kavli,irvington}
\icmlauthor{Daniel Muthukrishna}{kavli}
\icmlauthor{Michelle Lochner}{dopa,sarao}

\end{icmlauthorlist}

\icmlaffiliation{kavli}{Kavli Institute for Astrophysics and Space Research, Massachusetts Institute of Technology, Cambridge, MA 02139, USA}
\icmlaffiliation{irvington}{Irvington High School, 41800 Blacow Rd, Fremont, CA 94538, USA}
\icmlaffiliation{dopa}{Department of Physics and Astronomy, University of the Western Cape, Bellville, Cape Town, 7535, South Africa}
\icmlaffiliation{sarao}{South African Radio Astronomy Observatory, 2 Fir Street, Black River Park, Observatory, 7925, South Africa}

\icmlcorrespondingauthor{Daniel Muthukrishna}{danmuth@mit.edu}

\icmlkeywords{Machine Learning, ICML, Anomaly Detection, Transient, Astronomy}

\vskip 0.3in
]
\printAffiliationsAndNotice{}


\begin{abstract}

Automating anomaly detection is an open problem in many scientific fields, particularly in time-domain astronomy, where modern telescopes generate millions of alerts per night. Currently, most anomaly detection algorithms for astronomical time-series rely either on hand-crafted features or on features generated through unsupervised representation learning, coupled with standard anomaly detection algorithms. In this work, we introduce a novel approach that leverages the latent space of a neural network classifier for anomaly detection. We then propose a new method called Multi-Class Isolation Forests (MCIF), which trains separate isolation forests for each class to derive an anomaly score for an object based on its latent space representation. This approach significantly outperforms a standard isolation forest when distinct clusters exist in the latent space. Using a simulated dataset emulating the Zwicky Transient Facility (54 anomalies and 12,040 common), our anomaly detection pipeline discovered $46\pm3$ anomalies ($\sim 85\%$ recall) after following up the top 2,000 ($\sim 15\%$) ranked objects. Furthermore, our classifier-based approach outperforms or approaches the performance of other state-of-the-art anomaly detection pipelines when applied to the dataset used in \citet{Perez-Carrasco_2023}. Our novel method demonstrates that existing and new classifiers can be effectively repurposed for real-time anomaly detection. The code used in this work, including a Python package, is \href{https://github.com/Rithwik-G/AstroMCAD}{publicly available}.

% Anomaly detection is an open problem in many scientific fields, and automating it is imperative. Especially in time-domain astronomy, where modern telescopes generate thousands of alerts per night, it is often impossible for human experts to manually examine every new observation to find extremely rare anomalies. Currently, most anomaly detection algorithms for astronomical time-series rely either on hand-crafted features or on features generated through unsupervised representation learning, coupled with standard anomaly detection algorithms. In this work, we introduce an alternative approach: using the penultimate layer of a neural network classifier as the latent space for anomaly detection. We then propose a novel method, named Multi-Class Isolation Forests (\texttt{MCIF}), which trains separate isolation forests for each class to derive an anomaly score for an object from its latent space representation. This approach significantly outperforms a standard isolation forest. Using a simulated dataset emulating the Zwicky Transient Facility (54 anomalies and 12,040 common), our method discovered $46\pm3$ anomalies ($\sim 85\%$ recall) after following up the top 2,000 ($\sim 15\%$) ranked objects. Our novel method shows that existing and new classifiers can be effectively repurposed for real-time anomaly detection. The code used in this work (including a python package) is \href{https://github.com/Rithwik-G/AstroMCAD}{publicly available}.

\end{abstract}

\section{Introduction}
\label{sec:introduction}
Astronomical surveys measure light (or flux) in specific regions of the night sky. In time-domain astronomy, observations are made periodically, forming a light curve that represents the object's brightness variations over time. Most light curves exhibit minimal or gradual changes and are relatively unremarkable. However, when a significant deviation in brightness is detected with a high signal-to-noise ratio (S/N), it indicates the presence of a transient event in the observed galaxy. Transient events encompass a wide range of astrophysical phenomena, including various types of supernovae, which are explosive endings of stellar life cycles, and rare occurrences such as microlensing, where the light from a distant source is gravitationally amplified by an intervening massive object. Examples of light curves exhibiting transient events are presented in Appendix B.

With the advancement of these survey telescopes and the advent of large-scale transient surveys, we are entering a new paradigm for astronomical study. The Vera Rubin Observatory's Legacy Survey of Space and Time (LSST) is expected to observe ten million transient alerts per night \citep{Ivezic2009LSST:Products}. The traditional approach of manual examination of astronomical data, which has led to some of the biggest discoveries in astronomy, is no longer feasible. As a result, there is a growing need to develop methods that can automate the serendipity that has so far played a pivotal role in scientific discovery.

The literature on anomaly detection for astronomical transients presents two distinct problem definitions. Some approaches, categorized as unsupervised methods, focus on extracting anomalies from large datasets without relying on prior information \citep[e.g.][]{vraenn, Webb2020, Giles2019_Timeseries}. Numerous differing approaches exist for unsupervised anomaly detection. \citet{vraenn} used an unsupervised recurrent variational autoencoder to learn a representative latent space mapping of the light curves to then derive anomaly scores using an isolation forest. \cite{Webb2020} used a combination of a supervised clustering algorithm and an isolation forest.

In contrast, our work, among others \citep[e.g.][]{Perez-Carrasco_2023, OriginalPaper}, uses previous, either simulated or real, transients to determine whether a new light curve is anomalous. This approach is often referred to as novelty detection or supervised anomaly detection. Previous novelty detection approaches \citep[e.g.][]{OriginalPaper, Soraism2020Novelties} are often variations of one-class classification \citep{OneClassDef}. One-class classifiers attempt to model a set of \textit{normal} samples and then classify new transients as either part of that sample or as outliers. One-class methods have been shown to be effective at anomaly detection \citep{OneClass}, but they do not capture the complexity of the population of known astronomical transients, that are grouped into numerous classes with intrinsically different qualities. \citet{Perez-Carrasco_2023} extended the one-class classifier to multiple classes after training on features extracted from full light curve data. Their method adapts the single-class loss function to multiple classes by encouraging light curves of the same class to cluster together.

In this work, we leverage a light-curve classifier to address the one-class challenge and distinguish between the various classes of transients. Our approach demonstrates promising clustering in the feature space, the penultimate layer of the classifier, and shows a substantial level of discrimination in anomaly scores. Notably, similar feature extraction methods have shown potential in the field of astronomical image analysis \citep[e.g.][]{etsebeth2023astronomaly, WalmsleyCNN}.

Once a feature space has been identified using one of the previously mentioned methods, several prior works have employed an isolation forest \citep{isolationforest} to generate anomaly scores. While this approach has demonstrated success in previous research \citep[e.g][]{vraenn, Ishida2021_Timeseries, IsoForestUse2}, it faces challenges when dealing with a complex latent space that contains multiple clusters of intrinsically different transient classes. Consequently, the application of a single isolation forest may have limitations in accurately identifying certain anomalies, as it may struggle to adequately capture the distinct properties of each cluster. \citet{singh2022multiclass} also recognized the problem of using a single anomaly detector in a multi-class setting and introduced a method training an autoencoder for each class to then derive an anomaly score.

In response to this limitation, we propose the use of Multi-Class Isolation Forests (\texttt{MCIF}): a method that involves training a separate isolation forest for each known class and extracting the minimum score among them as the final anomaly score for a given sample. Our experimental results suggest that \texttt{MCIF} holds promise in improving anomaly detection performance for astronomical transients when there are defined clusters in the latent space.

% In summary, the main contributions of this work are:
% \begin{enumerate}
% \item The usage of the penultimate layer of a DNN classifier as the latent space in a multi-class anomaly detection setting. We show that an effective classifier can be repurposed for anomaly detection.
% \item We propose Multi-Class Isolation Forests for anomaly detection, an approach that trains an isolation forest for each class. This method is well-suited for the multi-class setting, and shows improvement from just an isolation forest when classes are well-defined.
% \end{enumerate}

\section{Dataset}
\label{sec:data_desc}
In this work, we use a collection of simulated light curves that match the observing properties of the Zwicky Transient Facility \citep[ZTF,][]{ZTF}. This dataset is described in \S~2 of \citet{OriginalPaper} and is based on the simulations developed for PLAsTiCC \citep{PlasticcSim}. Each transient in the dataset has flux and flux error measurements in the $g$ and $r$ passbands (two different light filters) with a median cadence of roughly 3 days in each passband.

The 17 transient classes we consider in this work are SNIa, SNIa-91bg, SNIax, SNIb, NIc, SNIc-BL, SNII, SNIIb, SNIIn, SLSN-I, PISN, KNe, AGN, TDE, ILOT, CaRT, and uLens-BSR. Due to their low occurrence in nature, \textbf{KNe, ILOT, CaRT, PISN, and uLens-BSR} are considered the \textbf{anomalous classes} in this work, and all remaining classes are considered the ``common'' classes. Example light curves from each of these classes are illustrated in Appendix \ref{sec:dataset}.

To emulate the real world, where scientists do not necessarily know what anomalies they are looking for, we ensure all transients from the anomalous classes are unseen by our model until final evaluation. Further, the goal of this work is to detect anomalies in general, not specifically transients of the aforementioned anomalous classes. Hence, we do not use physical priors of any transient type to aid in detection. Finally, because anomalies are inherently rare, but our simulated dataset is relatively class balanced, we perform evaluation by down-sampling the objects in the anomalous classes to create a more realistic evaluation dataset.

\section{Methods}

\subsection{Overview}

\begin{figure*}
    \centering
    \includegraphics[width=0.8\textwidth]{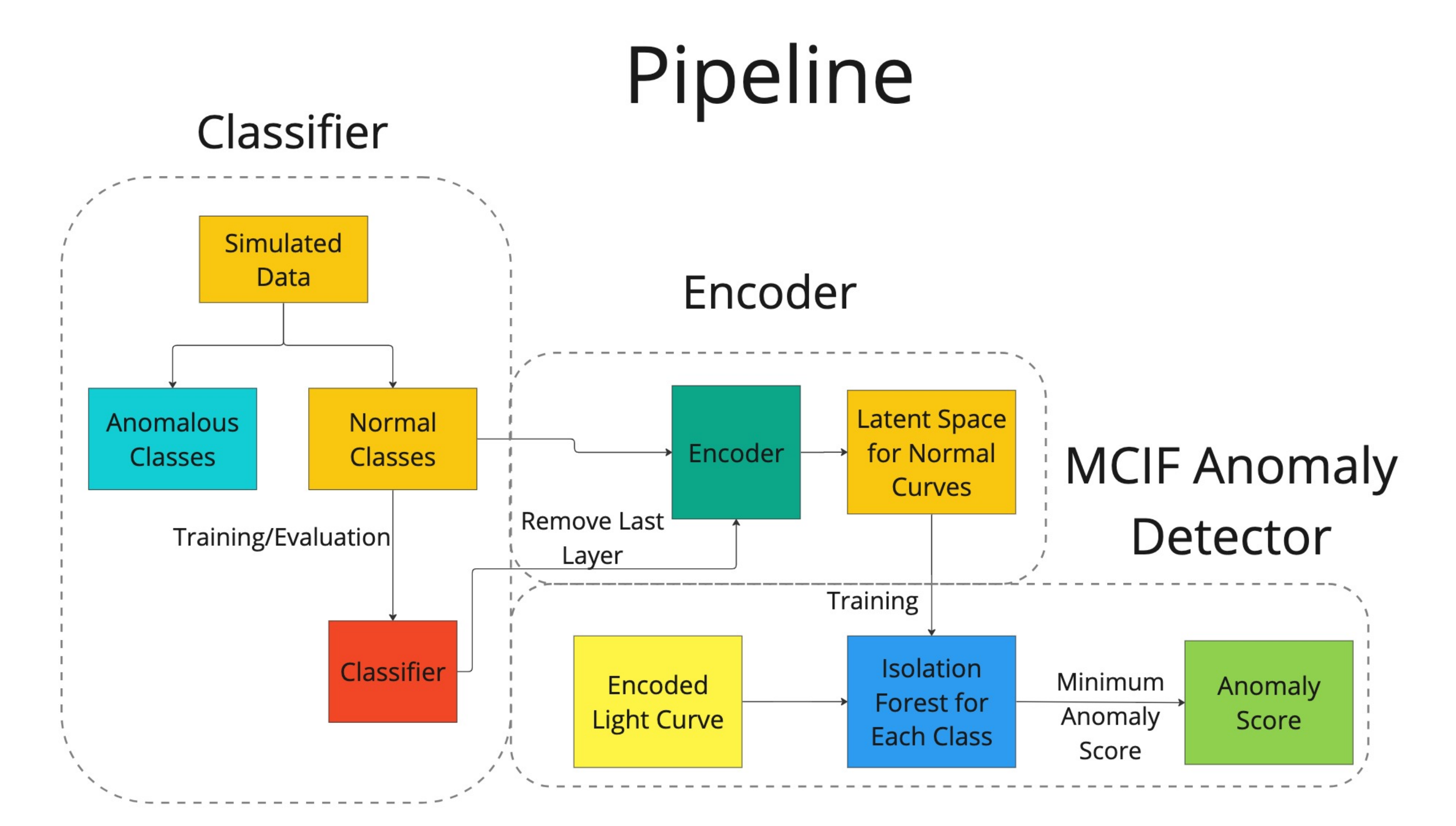}
    \caption{A visual summary of the architecture described in this work. Our approach first trains a classifier, then repurposes it as an encoder, and finally applies Multi-Class Isolation Forests (\texttt{MCIF}), proposed in this work, for anomaly detection.}
    \label{fig:Flowchart}
\end{figure*}

Figure \ref{fig:Flowchart} summarizes our methodology. We begin by training a Recurrent Neural Network (RNN) to classify common transient classes. After training, we remove the final classification layer of the model, repurposing the remaining architecture as an encoder. For effective anomaly detection, it's essential that transients from similar classes cluster together in this latent space. Our approach leverages the fact that this latent space was used for classification during training, naturally encouraging such clustering. 

Once we have established this representation space, we must extract anomalies from it. However, when dealing with multiple clusters, a single isolation forest may struggle to capture each cluster equally (for further details, refer to Section \ref{sec:MCIF_Advantages}). This challenge motivated our approach, \texttt{MCIF}, where we train an isolation forest for each class, representing a distinct cluster, and select the minimum anomaly score as the final score. This minimum score should come from the cluster to which the latent observation is closest, providing the desired functionality.

\subsection{Classifier}

We train a DNN (Deep Neural Network) classifier that maps a matrix of multi-passband light-curve data $\bm{X}_s$ for a transient $s$ to a $1 \times N_c$ vector of probabilities, reflecting the likelihood of the given light curve being from each of the aforementioned non-anomalous transient classes, where $N_c$ is the number of classes.

The transient classifier utilizes a Recurrent Neural Network (RNN) with Gated Recurrent Units \citep[GRU,][]{GRU} to handle the sequential time series data. GRUs have been shown to perform better than typical Recurrent Neural Networks (RNNs), have quicker training times than LSTMs\footnote{We empirically find that there is little difference between an LSTM and GRU model, in both classification accuracy and anomaly detection.} \citep{GRUvsLSTM}, and have shown promise in the domain of astronomical time-series \citep{OriginalPaper}. The input for each transient, $\bm{X}_s$, is a $4 \times N_T$ matrix where $N_T$ is the maximum number of timesteps for any input sample. $N_T$ is $656$ in this work, but most transients have much fewer observations. Each row of the input matrix is composed of the following vector,
\begin{equation}
     \bm{X}_{sj} = [f_{sj}, \epsilon_{sj}, t_{sj}, \lambda_p],
\end{equation}
where $f_{sj}$ is the scaled flux for the $j$th observation of transient $s$, $\epsilon_{sj}$ is the corresponding scaled uncertainty, $t_{sj}$ is the scaled time of when the measurement was taken, and $\lambda_p$ is the central wavelength of the passband from which the measurement comes from. 

% \begin{figure*}
% \centering
% \includegraphics[width=0.7\textwidth]{images/Classifier.pdf}
% \caption{A visualization of the neural network classifier being used in this work. Our model has two input streams, one for real-time light curve data and the other for contextual information. The light curve data (first input stream) goes through multiple GRU layers and then a dense layer. The contextual information (second input stream) feeds through a dense layer. The final dense layers from both input streams are merged into a concatenate layer. We feed that to a 100-neuron dense layer that will serve as the latent space of the encoder. Finally, this dense layer feeds into the output layer which provides classification scores.}
% \label{fig:ClassifierOutline}
% \end{figure*}

After the recurrent layers of the DNN, we pass some contextual information into the classifier, which has been shown to be helpful for light curve classification \citep{HostGalInfo}. In this work, we use the Milky Way extinction and the host galaxy's spectroscopic redshift as additional inputs to the network. We train our neural network for 40 epochs using the \texttt{adam} optimizer and counteract class imbalance in our dataset by using class weights inversely proportional to the frequency of the class while training. Our model takes roughly 10 minutes to train on a 16GB Tesla V100 GPU core.

One of the advantages of using a neural network-based architecture over hand-selected features is that it is a data-driven model, which should make it more sensitive to identifying out-of-distribution data. This inherent quality of neural networks makes them especially good for anomaly detection. However, the lack of interpretability of DNN models is a drawback and means that we can't discern why a certain object is marked anomalous.

\subsection{Anomaly Detection}

Once the classifier is trained, we remove the last layer and use the remaining architecture to map any light curve to the latent space. We define this encoder as a function $E(\bm{X}_s)$, that takes the aforementioned preprocessed light curve data, $\bm{X}_s$, and maps it to a 100-dimensional latent space $\bm{z}_s$

\begin{equation}
    \bm{z}_s = E(\bm{X}_s)
\end{equation}

For anomaly detection, we now want to compute the anomaly score, $a_s = A(\bm{z}_s)$, where $A(\bm{z}_s)$ is a function that evaluates the anomaly score $a_s$ for a latent observation $\bm{z}_s$. The goal of this work is to generate relatively large anomaly scores for anomalous transients and smaller anomaly scores for non-anomalous transients. 

% Isolation Forests are known to be a very simple yet effective anomaly detection algorithm, especially in the domain of astronomical time series. However, using a single isolation forest performs poorly in determining some common classes as non-anomalous. 
We propose a new framework where an isolation forest is trained separately on data from every class, using the minimum anomaly score from any isolation forest as the final anomaly score\footnote{We also tested using an SVM and the distance from the cluster's center, but an Isolation Forest empirically worked the best, as is seen in similar literature.}. We call this approach Multi-Class Isolation Forests (\texttt{MCIF}).

We define $12$ isolation forests, $I_c(\bm{z}_s)$, trained on latent space observations from the common transient class $c$. The final anomaly score is defined as

\begin{equation}
    A(\bm{z}_s) = \min_{\forall c} \Bigl( -I_c(\bm{z}_s) \Bigr)
\end{equation}

The function $I_c(\bm{z}_s)$ is positive for less anomalous transients and negative for anomalous ones, to be consistent with the \texttt{sklearn} implementation of Isolation Forests. We negate the scores as we prefer defining transients with higher anomaly scores to be more anomalous, but this makes no difference to the results. All isolation forests used in this work are trained with 200 estimators. The results of using a single isolation forest and the benefits of using Multi-Class Isolation Forests are explored further in Section \ref{sec:MCIF_Advantages}.

% \subsection{Hyperparameter Tuning}

% Anomaly detection is unique in the regard that there is no way to evaluate an anomaly detection pipeline before final testing, when the anomalous data is unveiled. Hence, we do not tune hyperparameters for model selection and instead retrospectively analyze effects of choosing different hyperparameters. The effects of scaling the latent space size on anomaly detector performance are discussed in Section \ref{sec:hyperparameter}.
% The classifier trained on our data is not tuned in this manner as it would be computationally expensive on our considerably larger dataset, and that we want to show the ability of a classifier not necessarily intended for anomaly detection.

% Despite this challenge, we train models with different hyperparameters in Appendix [WORK IN PROGRESS].

\section{Evaluation}

\subsection{Latent Space}

\begin{figure*}
\centering
\begin{tabular}{ll}
\includegraphics[width=0.45\textwidth]{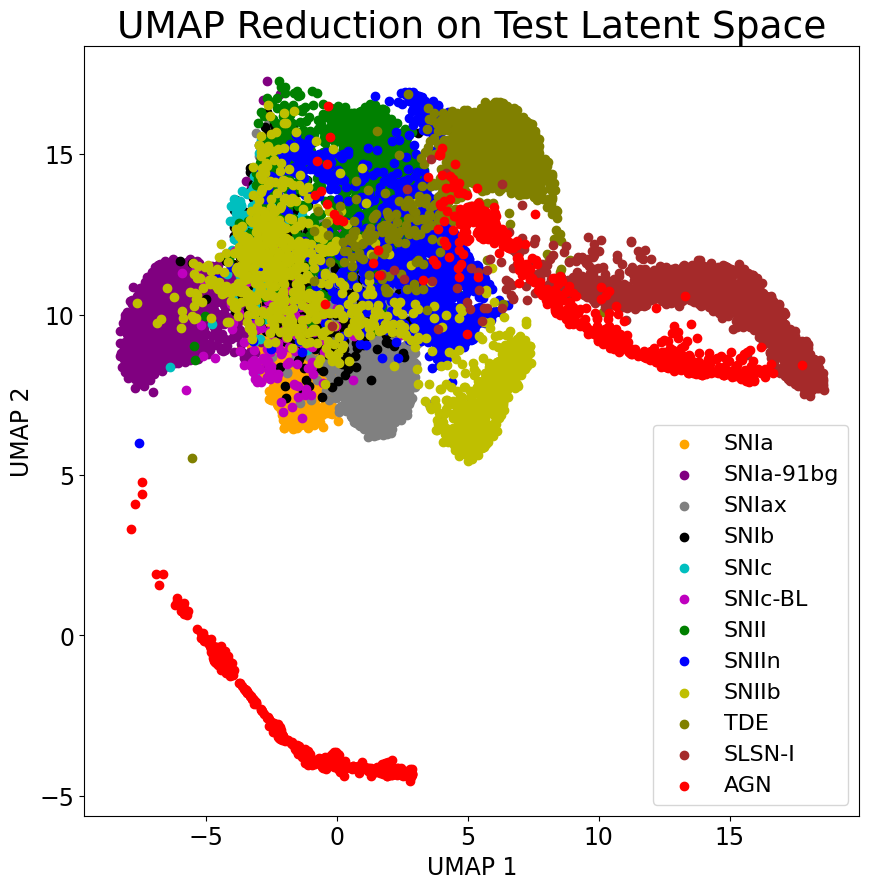}
&
\includegraphics[width=0.45\textwidth]{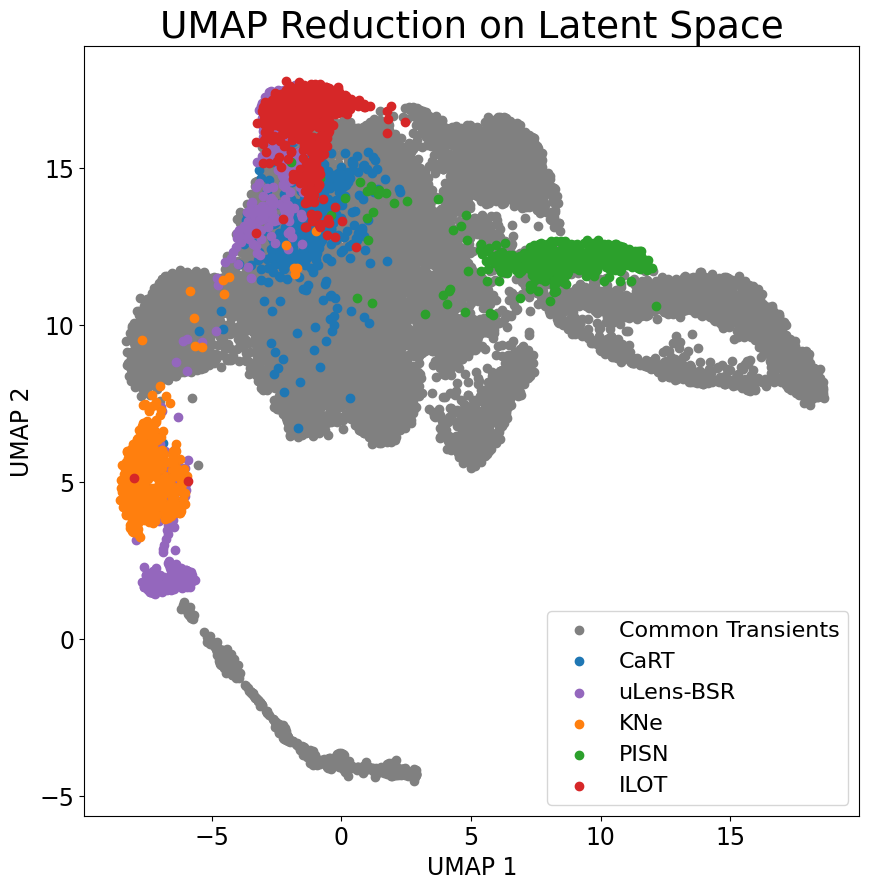}
\end{tabular}
\caption{The UMAP reduction of the latent space derived from the test set, which includes 10\% of the common transients reserved for testing the classifier [left] and randomly sampled anomalous transients from the unseen anomaly dataset [right]. Despite not being trained on this data, the learned features still exhibit clear visual structure and anomalous transients form distinct clusters separate from the common classes. It is important to note that the UMAP reduction is used only for visualization purposes, and the actual anomaly detection is performed on the nine-dimensional latent space.}
\label{fig:UMAP}
\end{figure*}

After repurposing the classifier as an encoder, we obtain a 100-dimensional latent space. We can visualize this latent space with UMAP \citep{UMAP}, a manifold embedding technique, to determine if there is visible clustering\footnote{We use the \texttt{umap-learn} implementation in \texttt{python} using the hyperparameters ``minimum distance'' set to 0.5 and ``number of neighbors" set to 500.}. In Figure \ref{fig:UMAP} [left], we plot the UMAP representations of the test data. While it is difficult to examine some of the overlapping classes in this embedded space, there is clear clustering of many of the classes.
In Figure \ref{fig:UMAP} [right], we color all of the common classes grey and include a sample of transients from the anomalous classes. We see that the anomalous classes cluster together in the embedded space and separate from the common transients despite the model not being trained on these objects. This level of clustering suggests that our encoder may be discovering generalizable patterns within light curves, and this property may have potential use cases beyond anomaly detection in few-shot classification.
It is important to note that we only use UMAP for visualisation purposes and that the latent space used for anomaly detection is obtained directly from the penultimate layer of the classifier.

\subsection{Anomaly Detection}

\begin{figure*}[h]
\centering
\begin{tabular}{ll}
\includegraphics[width=0.45\textwidth]{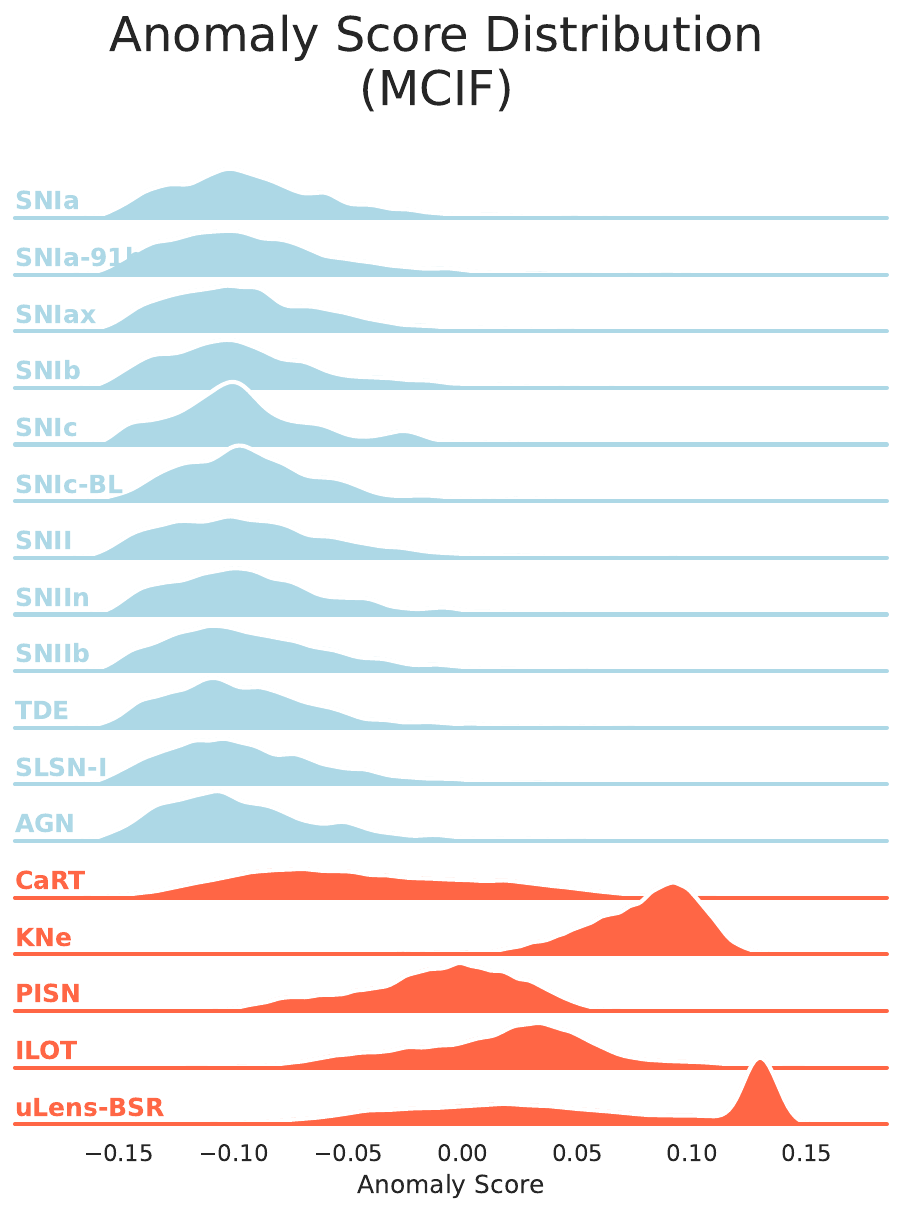}
&
\includegraphics[width=0.45\textwidth]{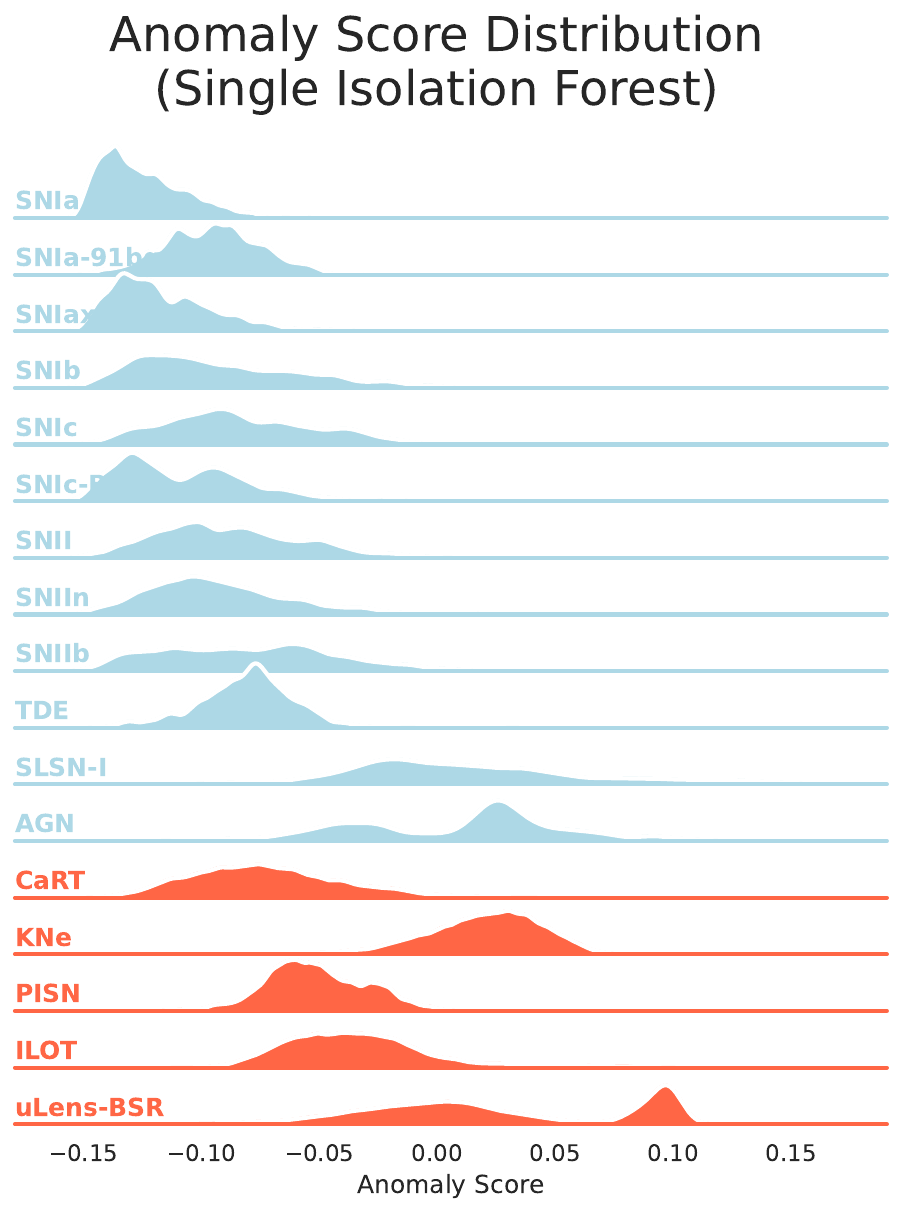}
\end{tabular}
\caption{The distribution of anomaly scores for each class, computed using \texttt{MCIF} [left] or a single isolation forest [right] on the latent representations derived from full light curves. The scores are plotted using $100\%$ of the anomalous dataset (unseen during training) and the test dataset of common classes. The anomalous classes (bottom five in red) generally show higher anomaly scores with positively skewed distributions when using \texttt{MCIF}, however this is less true when using a single isolation forest. The common classes and CaRTs all have low anomaly scores when using \texttt{MCIF}.}
\label{fig:Distribution}
\end{figure*}

In Figure \ref{fig:Distribution} [left], we plot the distribution of anomaly scores predicted by \texttt{MCIF} from the latent space for each class. The plot demonstrates the distinction in anomaly scores of common and anomalous transients as there is a significant skew towards larger anomaly scores for the anomalous classes. However, Calcium Rich Transients (CaRTs), despite being one of our anomalous classes, tend to have lower anomaly scores. CaRTs are notoriously difficult to photometrically classify as anomalous due to their resemblance to other common supernova classes (see Fig. 8 of \citealt{Muthukrishna19RAPID} for example).

\begin{figure*}
  \centering
  \begin{tabular}{ll}
  \includegraphics[scale=0.5]{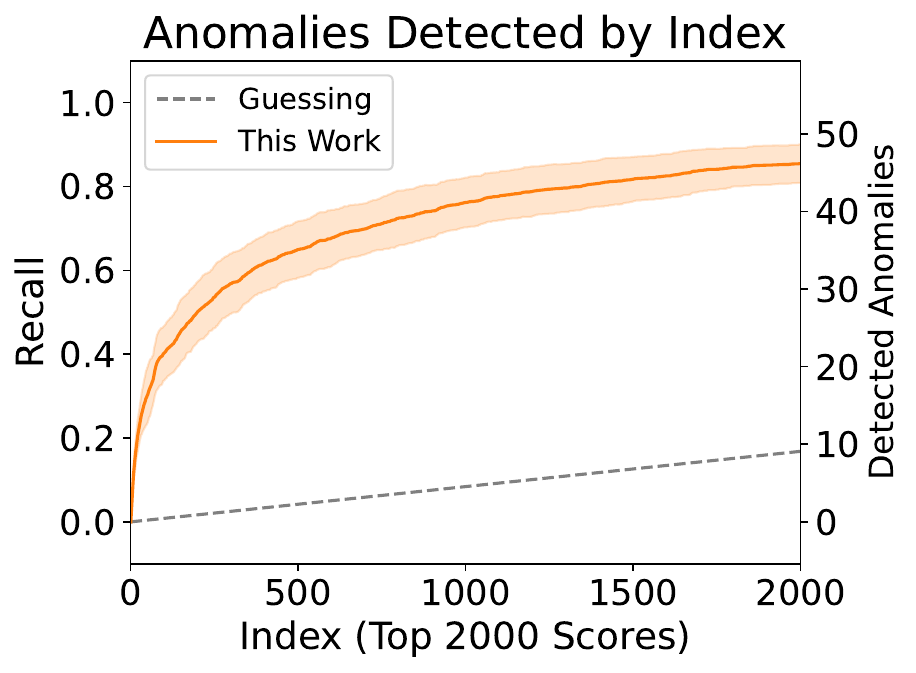}  
  &
  \includegraphics[scale=0.5]{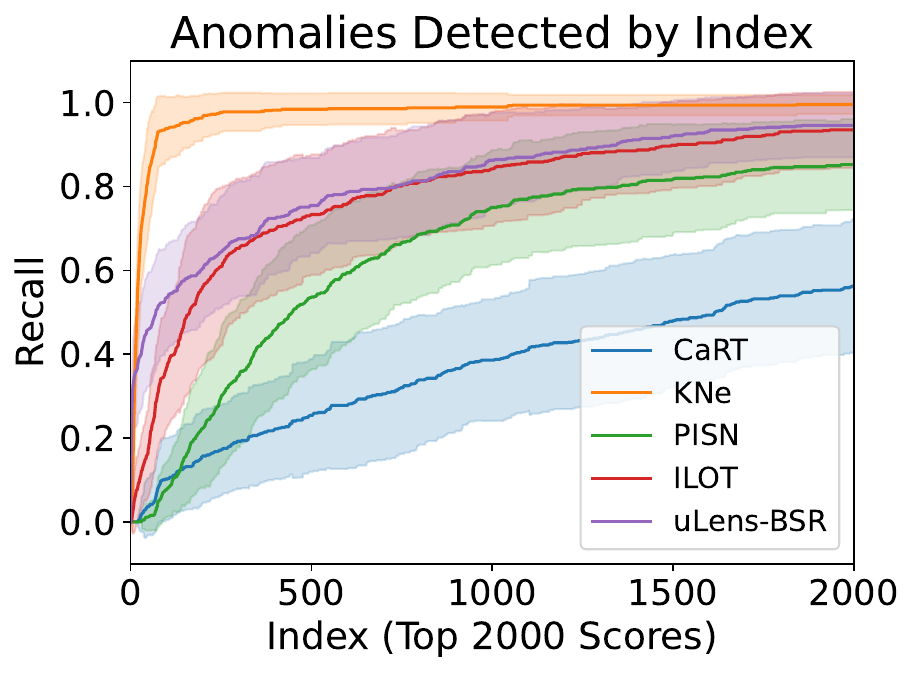}  
  \end{tabular}

  \caption{Anomalies detected in the 2,000 top-ranked transients by \texttt{MCIF} anomaly score index, using a test sample reflecting the estimated frequency of anomalies in nature. In the sample of 12,040 common transients and 54 anomalous transients, the model recalls $46\pm3$ $(\sim85\%)$ of the anomalies after following up the top 2,000 ranked transients. The left plot aggregates all anomalies and the right plot delineates per class.
  To control for the variance imposed by the small anomaly sample size, we repeat the sampling 50 times. The mean and standard deviation of detected anomalies are plotted as the solid lines and shaded regions, respectively. }
  \label{fig:Index}
\end{figure*}

\subsection{Detection Rates in a Representative Population}

\label{sec:representative_population}
The previous results do not acknowledge a key difficulty of anomaly detection: anomalies are inherently infrequent. While the frequency of anomalous transients in nature is not known, a good estimate for the expected population frequency was presented in \citet{PlasticcSim} for the PLAsTiCC dataset \citep{PlasticcData}. Using PLAsTiCC frequencies for each class, the rate of common transients is roughly 220 times that of anomalous transients. We used this rate to randomly select a more realistic test dataset that contained 12,040 normal transients and 54 anomalies. Randomly selecting a representative sample of only 54 anomalies is subject to significant variance. Therefore, we created 50 sample datasets to perform 50-fold cross-validation. Information on the exact composition of these test sets is listed in Table \ref{table:populationdistribution}.

For each validation set, we ranked the transients by the anomaly scores predicted by \texttt{MCIF}. We then selected the top 2,000 highest-scoring transients (roughly 15\% of the dataset) as the candidate pool. Across 50 repeated trials, we identified $46\pm3$ out of the 54 true anomalies in our dataset (recalling $\sim85\%$ of the anomalies). In Figure \ref{fig:Index}, we plot the fraction of anomalies recalled and the total number of anomalies recovered for thresholds up to the top 2,000 transients. \texttt{MCIF} recalls the majority of true anomalies among candidates having the highest anomaly scores, followed by a tapering as fewer anomalies remain.

\subsection{Comparison Against Other Approaches}
\label{sec:benchmarking}

\begin{table*}[!ht]
  
  \hskip-3.3cm
  \begin{center}
  \scalebox{0.60}{
  \begin{tabular}{|l||cccc||ccccc||ccccc|} 
    \hline 
    & \multicolumn{4}{c||}{ \textbf{Transient}}&
    \multicolumn{5}{c||}{ \textbf{Stochastic}}&
    \multicolumn{5}{c|}{ \textbf{Periodic}} \\

    Method  & SLSN & SNII & SNIa & SNIbc & AGN & Blazar & CV/Nova & QSO & YSO & CEP & DSCT & E & RRL & LPV \\ \hline
    
    IForest  & $0.640$ & $0.721$ & $0.428$ & $0.490$ & $0.573$ & $0.710$ & $\mathbf{0.975}$ & $0.468$ & $\mathbf{0.913}$ & $0.359$ & $0.295$ & $0.469$ & $0.549$ & $\mathbf{0.971}$ \\
    \citep{isolationforest}
    & $\pm 0.014$ & $\pm 0.021$ & $\pm 0.032$ & $\pm 0.038$ & $\pm 0.017$ & $\pm 0.009$ & $\mathbf{\pm 0.001}$ & $\pm 0.016$ & $\mathbf{\pm 0.003}$ & $\pm 0.007$ & $\pm 0.012$ & $\pm 0.021$ & $\pm 0.033$ & $\mathbf{\pm 0.007}$ \\ \hline
  
    OCSVM  & $0.577$ & $0.587$ & $0.434$ & $0.492$ & $0.532$ & $0.443$ & $0.909$ & $\mathbf{0.517}$ & $0.792$ & $0.432$ & $\mathbf{0.557}$ & $0.555$ & $0.539$ & $0.943$ \\ \citep{OneClassDef}
    & $\pm 0.014$ & $\pm 0.014$ & $\pm 0.021$ & $\pm 0.011$ & $\pm 0.008$ & $\pm 0.002$ & $\pm 0.001$ & $\mathbf{\pm 0.005}$ & $\pm 0.005$ & $\pm 0.004$ & $\mathbf{\pm 0.005}$ & $\pm 0.003$ & $\pm 0.004$ & $\pm 0.001$ \\ \hline
    
    AE  & $\mathbf{0.736}$ & $\mathbf{0.807}$ & $0.438$ & $0.537$ & $\mathbf{0.701}$ & $\mathbf{0.762}$ & $\mathbf{0.980}$ & $0.443$ & $\mathbf{0.990}$ & $0.564$ & $0.367$ & $\mathbf{0.864}$ & $\mathbf{0.907}$ & $\mathbf{0.996}$ \\
    \citep{rumelhart_1987}
     & $\mathbf{\pm 0.022}$ & $\mathbf{\pm 0.021}$ & $\pm 0.015$ & $\pm 0.019$ & $\mathbf{\pm 0.010}$ & $\mathbf{\pm 0.006}$ & $\mathbf{\pm 0.016}$ & $\pm 0.004$ & $\mathbf{\pm 0.001}$ & $\pm 0.024$ & $\pm 0.015$ & $\mathbf{\pm 0.009}$ & $\mathbf{\pm 0.015}$ & $\mathbf{\pm 0.000}$ \\ \hline
    
    VAE  & $0.669$ & $0.690$ & $0.404$ & $0.522$ & $0.596$ & $0.597$ & $0.849$ & $\mathbf{0.500}$ & $0.795$ & $0.442$ & $0.417$ & $0.561$ & $0.451$ & $0.936$ \\
    \citep{kingma_2013}
     & $\pm 0.015$ & $\pm 0.023$ & $\pm 0.018$ & $\pm 0.025$ & $\pm 0.007$ & $\pm 0.010$ & $\pm 0.028$ & $\mathbf{\pm 0.009}$ & $\pm 0.009$ & $\pm 0.010$ & $\pm 0.007$ & $\pm 0.007$ & $\pm 0.006$ & $\pm 0.007$ \\ \hline
    
    Deep SVDD  & $0.644$ & $0.731$ & $0.475$ & $0.507$ & $0.496$ & $0.607$ & $0.932$ & $0.411$ & $0.901$ & $0.707$ & $0.482$ & $0.636$ & $0.774$ & $0.785$ \\
    \citep{ruff_2018}
    & $\pm 0.043$ & $\pm 0.043$ & $\pm 0.040$ & $\pm 0.040$ & $\pm 0.025$ & $\pm 0.044$ & $\pm 0.015$ & $\pm 0.008$ & $\pm 0.022$ & $\pm 0.027$ & $\pm 0.054$ & $\pm 0.055$ & $\pm 0.068$ & $\pm 0.025$ \\ \hline
    
    MCDSVDD  & $\mathbf{0.686}$ & $\mathbf{0.828}$ & $\mathbf{0.624}$ & $\mathbf{0.584}$ & $\mathbf{0.706}$ & $0.512$ & $0.770$ & $0.483$ & $0.854$ & $\mathbf{0.858}$ & $\mathbf{0.819}$ & $\mathbf{0.945}$ & $\mathbf{0.953}$ & $0.953$ \\
    \citep{Perez-Carrasco_2023}
    & $\mathbf{\pm 0.051}$ & $\mathbf{\pm 0.024}$ & $\mathbf{\pm 0.039}$ & $\mathbf{\pm 0.032}$ & $\mathbf{\pm 0.069}$ & $\pm 0.113$ & $\pm 0.127$ & $\pm 0.080$ & $\pm 0.041$ & $\mathbf{\pm 0.025}$ & $\mathbf{\pm 0.015}$ & $\mathbf{\pm 0.006}$ & $\mathbf{\pm 0.003}$ & $\pm 0.008$ \\ \hline

    Classifier + IForest & $\mathbf{0.757}$ & $\mathbf{0.811}$ & $\mathbf{0.619}$ & $0.556$ & $\mathbf{0.715}$ & $\mathbf{0.720}$ & $0.945$ & $0.456$ & $\mathbf{0.977}$ & $\mathbf{0.766}$ & $0.504$ & $\mathbf{0.811}$ & $\mathbf{0.907}$ & $\mathbf{0.969}$ \\
(This work) & $\mathbf{\pm0.047}$ & $\mathbf{\pm0.017}$ & $\mathbf{\pm0.073}$ & $\pm0.039$ & $\mathbf{\pm0.028}$ & $\mathbf{\pm0.032}$ & $\pm0.015$ & $\pm0.041$ & $\mathbf{\pm0.003}$ & $\mathbf{\pm0.066}$ & $\pm0.111$ & $\mathbf{\pm0.038}$ & $\mathbf{\pm0.026}$ & $\mathbf{\pm0.016}$ \\ \hline

    % Classifier + MCIF & $0.577$ & $0.754$ & $\mathbf{0.583}$ & $\mathbf{0.539}$ & $0.594$ & $0.695$ & $0.901$ & $\mathbf{0.546}$ & $0.908$ & $\mathbf{0.809}$ & $\mathbf{0.700}$ & $0.771$ & $0.785$ & $0.862$ \\ 
    % (This work) & $\pm0.128$ & $\pm0.061$ & $\mathbf{\pm0.075}$ & $\mathbf{\pm0.044}$ & $\pm0.119$ & $\pm0.04$ & $\pm0.031$ & $\mathbf{\pm0.097}$ & $\pm0.039$ & $\mathbf{\pm0.085}$ & $\mathbf{\pm0.014}$ & $\pm0.021$ & $\pm0.09$ & $\pm0.07$ \\ \hline

    Classifier + MCIF & $0.567$ & $0.699$ & $\mathbf{0.536}$ & $\mathbf{0.560}$ & $0.615$ & $0.701$ & $0.882$ & $\mathbf{0.605}$ & $0.893$ & $\mathbf{0.875}$ & $\mathbf{0.742}$ & $0.773$ & $0.808$ & $0.779$ \\
    (This work) & $\pm0.091$ & $\pm0.046$ & $\mathbf{\pm0.061}$ & $\mathbf{\pm0.034}$ & $\pm0.048$ & $\pm0.045$ & $\pm0.050$ & $\mathbf{\pm0.051}$ & $\pm0.025$ & $\mathbf{\pm0.036}$ & $\mathbf{\pm0.044}$ & $\pm0.031$ & $\pm0.046$ & $\pm0.107$ \\ \hline

    MCIF & $0.503$ & $0.668$ & $0.532$ & $\mathbf{0.643}$ & $0.614$ & $\mathbf{0.745}$ & $\mathbf{0.966}$ & $0.446$ & $0.907$ & $0.514$ & $0.433$ & $0.476$ & $0.447$ & $0.959$ \\  
    (This work) & $\pm0.018$ & $\pm0.008$ & $\pm0.007$ & $\mathbf{\pm0.005}$ & $\pm0.02$ & $\mathbf{\pm0.008}$ & $\mathbf{\pm0.003}$ & $\pm0.007$ & $\pm0.007$ & $\pm0.013$ & $\pm0.009$ & $\pm0.021$ & $\pm0.011$ & $\pm0.004$ \\ \hline

  \end{tabular} }
  \end{center}
  
    \caption{Performance of each model when applied to the dataset used in \citet{Perez-Carrasco_2023}. Each row represents a different anomaly detection algorithm and each column is a different class being chosen as the anomalous class. The performance is evaluated using the AUROC score of detected anomalies. The top 3 metrics per class are marked in bold. The AUROC scores for the first 5 methods are taken directly from and are reported in \citet{Perez-Carrasco_2023}. A visual representation of this table is shown in Figure \ref{fig:visual_table}.}
  \label{table:results}
\end{table*}

In the field of anomaly detection in time-domain astronomy, there is no comprehensive baseline on which to evaluate different detection methods. This is largely because of the vastly differing definitions of what \textit{anomaly detection} is, for example, the difference between unsupervised and novelty detection methods as described in Section \ref{sec:introduction}. Baselining all existing anomaly detection methods is a much needed line of future work, especially as there is no consensus on which method will work best on the deluge of data that will available when LSST is running.

Despite these challenges, \citet{Perez-Carrasco_2023} evaluated 5 different approaches to anomaly detection (see Table \ref{table:results} for all benchmarked approaches), and we use their dataset (which was inspired by \citealt{SanchezSaez2021}) to evaluate our classifier-based approach. In contrast to our dataset of raw light curve data, this dataset consists of \textit{features} extracted from light curves. We evaluate three new techniques for anomaly detection on this dataset: using a classifier with \texttt{MCIF}, a classifier with just a single Isolation Forest, and \texttt{MCIF} on its own\footnote{We can use \texttt{MCIF} on its own as this is a dataset of features extracted from time-series, not the raw time-series.}. The dataset is split into 3 hierarchical categories with 4-5 transient classes each. Evaluation is performed separately for each class, each time counting that transient class as anomalous and the rest of its hierarchical category as common. Full evaluation is performed across 5 folds of testing data for cross-validation.

% For each fold of the 5-fold cross-validation performed for evaluation, we train multiple models and use the model with the highest silhouette score \citep{Rousseeuw1987}, a clusterability metric, on the training data for final model selection. We choose to do this as a well-clustered latent space is likely easier to extract anomalies from, and we find a correlation between the clusterability and anomaly detector performance empirically. Training using the silhouette score as the loss function is out of the scope of this work, but is a promising direction of future work.

As seen in Table \ref{table:results} (and visually in Figure \ref{fig:visual_table}), our classifier-based approach with an isolation forest is one of the top approaches for most transient classes, showing the power of using a classifier's latent space for anomaly detection. Using a classifier with \texttt{MCIF} also preforms promisingly, however is sometimes worse than using a classifier with a single isolation forest. This is not the case on our dataset and is discussed further in the next section.

% This contradicts the result shown in Section \ref{sec:MCIF_Advantages}, where \texttt{MCIF} improves results considerably. We find that \texttt{MCIF} improves results when clusters are more well-defined, and this usually comes with more data. Further analysis on when \texttt{MCIF} improves results is present in Appendix \ref{sec:comparison_analysis}.

% Our method's performance reasonably approaches the SOTA for this task and shows that existing classifiers can be used for anomaly detection, especially when training new models is expensive.

% analyze when mcif does better on the latent space and when it doesn't, using the UMAP plots
% argument is that you need a fairly good idea of what different classes are

% \texttt{MCIF} also performs well on this benchmark, and nears the best performing method for numerous classes. We theorize that when \texttt{MCIF} outperforms our classifier-based approach (which relies on \texttt{MCIF}), there is not enough data for a classifier create a representative latent space with reasonable clustering. This usually occurs for the stochastic transient classes. This is yet another example of no free lunch: being trained for a task other than anomaly detection, the model needs more data to generalize. The high performance deviations seen in the results of the classifier reaffirms that our classifier-based anomaly detection approach requires more data, as we find much less deviation when training with our larger, simulated dataset.

\subsection{Advantages of MCIF}

\label{sec:MCIF_Advantages}

% \begin{figure*}

% \centering

% \begin{tabular} {@{}c@{}}
% \includegraphics[width=0.8\textwidth]{images/MedianScoreRedo.pdf}
% \end{tabular}

% \begin{tabular} {@{}c@{}}
% \includegraphics[width=0.8\textwidth]{images/MedianScoreSingle.pdf}
% \end{tabular}

% \caption{The median anomaly score for each class computed for latent representations of transients obtained from full light curves when a single isolation forest is used for anomaly detection [bottom] and when \texttt{MCIF} is used [top]. The scores are derived from the unseen anomalous data and the common transient testing data. The right 5 classes (scores in bold) are anomalous. The common classes have somewhat lower median scores when using a single isolation forest, but the common classes SLSN-I and AGN (among others) are considered very anomalous, unlike when using \texttt{MCIF}.}
% \label{fig:FullAverageScore}
% \end{figure*}
To evaluate \texttt{MCIF}, we compare it to the performance of using a normal isolation forest to detect anomalies from the latent representation $z_s$ of a light curve\footnote{Note that this evaluation is done on the dataset described in Section \ref{sec:data_desc}, not the one used for comparitive analysis in Section \ref{sec:benchmarking}.}. We train an isolation forest on the latent represenation of our training data using $2400$ estimators (the same number used by all of the isolation forests in \texttt{MCIF} combined). To account for the class imbalance in our training data, we weight samples from underrepresented classes more heavily during the training of the isolation forest, using the same weighting scheme used in the classifier. The anomaly score function $A(z_s)$ is now simply the negated anomaly score output from a single isolation forest trained on all the latent representations of the training data.

As shown in Figure \ref{fig:Distribution} [right], there is little distinction in the anomaly scores of most anomalous and common classes when using a single isolation forest. Surprisingly, the common classes SLSN-I and AGN are classified as relatively more anomalous than all the other classes.

The UMAP reduction of the latent space of our classifier, as depicted in Figure \ref{fig:UMAP}, provides insight into this behaviour. The SLSN-I and AGN classes are located far from the main cluster formed by other classes and are nearly perfectly classified by our classifier (shown in the confusion matrix and ROC curves in Figure \ref{fig:ConfusionROC} in Appendix \ref{sec:classifier}). In fact, the near-perfect classification hinted at their potential to be misidentified as anomalies, suggesting that their distinct characteristics make them easily separable from other classes and, consequently, more likely to be flagged as anomalous by a single isolation forest. On the other hand, while SNIa also deviate from the central cluster in the UMAP visualization, they are among the most challenging classes to classify accurately and are the most frequently observed transient class in real surveys. Thus, they are likely a part of the central cluster in the full 100-dimensional latent space. 
Hence, while an isolation forest is good at detecting anomalies, it struggles to capture the structure of a latent space with numerous well-defined clusters. This drawback of using a single isolation forest could explain why other works report high anomaly scores for SLSN-I and AGN \citep[e.g.][]{vraenn}. Using a class-by-class (or cluster-by-cluster) anomaly detector, such as \texttt{MCIF}, can mitigate this. A direct comparison of the anomaly score distributions in Figure \ref{fig:Distribution} empirically demonstrates the advantages of \texttt{MCIF} on our dataset. 

\begin{figure*}
\centering
\begin{tabular}{ll}
\includegraphics[width=0.45\textwidth]{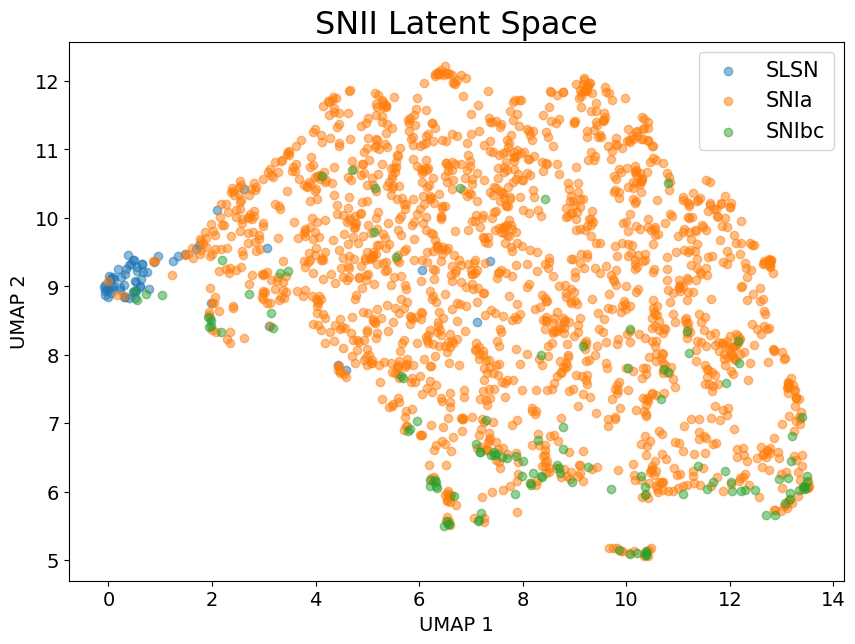}
&
\includegraphics[width=0.45\textwidth]{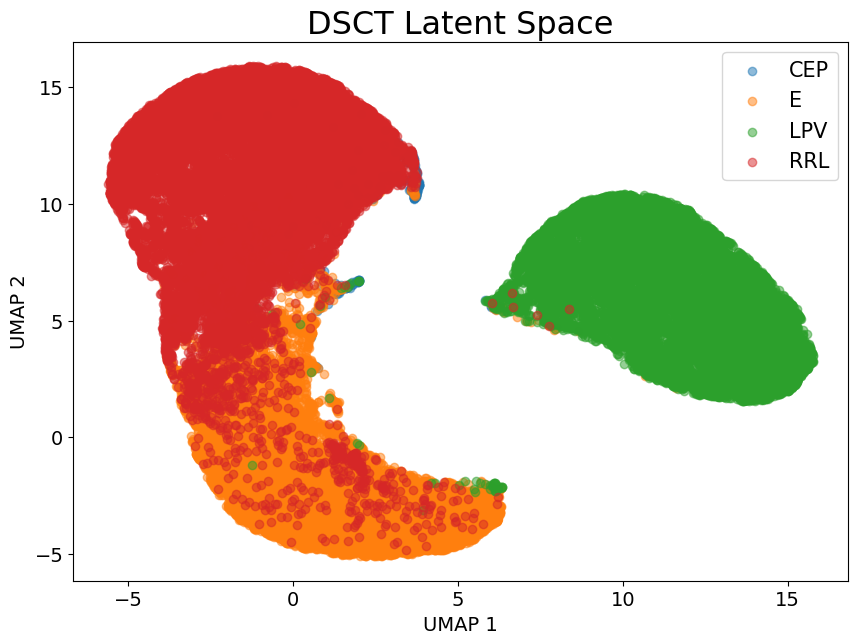}
\end{tabular}
\caption{The UMAP reduction of the training data in the latent space for a classifier trained for detecting the class SNII [left] and DSCT [right] as anomalous using the data introduced in \cite{Perez-Carrasco_2023} and used in Section \ref{sec:benchmarking}. As the UMAP only plots the training data, it includes all the classes in the respective hierarchical category (seen in Table \ref{table:results}) but the one set aside as anomalous. }
\label{fig:umapres}
\end{figure*}

Further analysis of MCIF's performance on the comparative evaluation dataset (Section \ref{sec:benchmarking}) reveals that, contrary to the results shown in Figure \ref{fig:Distribution}, a single isolation forest generally outperforms MCIF (Table \ref{table:results}). Investigating the UMAP representations of the latent space for classes exhibiting this discrepancy offers insights. When SNII is considered anomalous, the latent space (Figure \ref{fig:umapres} [left]) lacks clear separation between SNIbc and SNIa, likely due to poor generalization caused by the limited number of SNIbc transients in the training set, explaining the single isolation forest's superior performance. However, for the DSCT class (Figure \ref{fig:umapres} [right]), distinct visual clusters are present, and MCIF achieves better results. These findings suggest that MCIF enhances performance when majority classes are well-separated, a characteristic seemingly inherent to the dataset rather than the classifier-based latent space identification approach, as a single isolation forest surpasses MCIF on the raw data for most classes where it also outperforms MCIF on the classifier's latent space. Future research should explore the factors influencing MCIF's effectiveness based on the separability of raw data, with the SNII case indicating a partial dependence on data quantity, as increased data improves the DNN's generalization ability.

\subsection{Scaling the Latent Space}

Anomaly detection presents a unique challenge in terms of evaluation, as the true anomalies are only revealed during the final testing phase. Consequently, we refrain from tuning hyperparameters for model selection and instead retrospectively analyze the effects of different hyperparameter choices, particularly the size of the latent space.

\label{sec:hyperparameter}

\begin{figure}[h]
\centering
\includegraphics[width=0.45\textwidth]{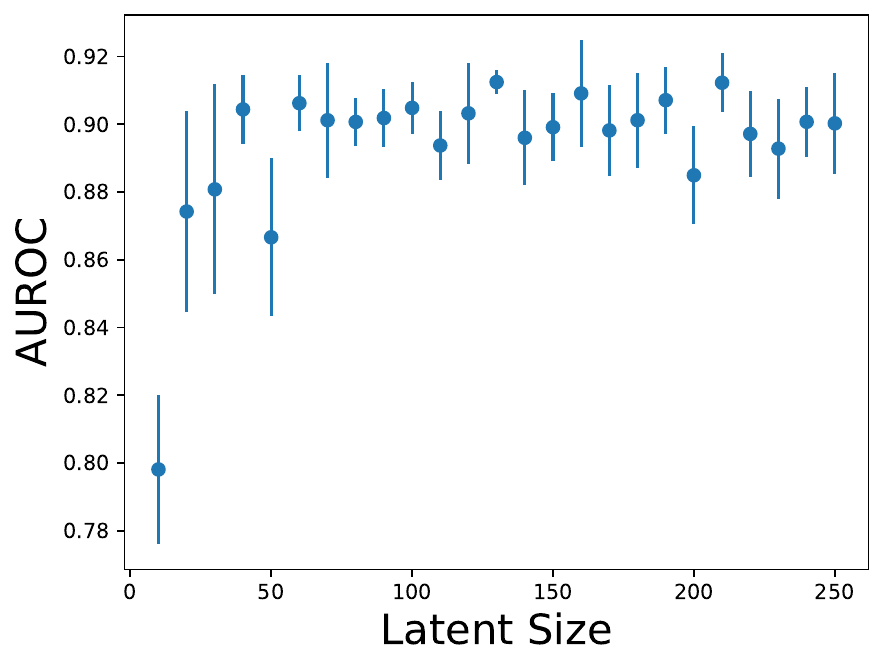}
\caption{Anomaly detection performance (AUROC) of models trained with different latent space sizes. A significant improvement is observed when increasing the latent size up to 50 dimensions, with performance plateauing thereafter.}
\label{fig:scaling}
\end{figure}

To assess the impact of latent space size on anomaly detection performance, we train multiple models with varying latent dimensions and evaluate them using the AUROC. As shown in Figure \ref{fig:scaling}, increasing the latent size beyond 50 leads to significant improvements in anomaly detection performance, with diminishing returns observed after 70 dimensions. Smaller models generally exhibit lower average performance and higher variance. Interestingly, we do not observe a performance drop in high-dimensional latent spaces, despite the presence of numerous correlated features. This robustness can be attributed to the ability of isolation forests and ensemble methods to effectively handle high-dimensional data. Our classifier's 100-neuron penultimate layer is one of the best-performing hyperparameter settings, with any reasonably large latent space yielding comparable results.

It is worth noting that while classifiers demonstrate effectiveness in anomaly detection, we find little correlation between classification accuracy and anomaly detection performance. This highlights a key drawback in terms of interpretability in both DNN frameworks and our approach, warranting further investigation.

\section{Conclusion}

In this work, we have introduced a novel approach that leverages the latent space of a neural network classifier for identifying anomalous transients. Our pipeline, which combines a deep recurrent neural network classifier with our novel Multi-Class Isolation Forest (\texttt{MCIF}) anomaly detection method, demonstrates promising performance on simulated data matched to the characteristics of the Zwicky Transient Facility and when compared to other state-of-the-art anomaly detection methods.

The key advantages of our approach are: 
\begin{enumerate}

\item The recurrent neural network (RNN) classifier maps light curves into a low-dimensional latent space that naturally clusters similar transient classes together, providing an effective representation for anomaly detection. We repurposed the penultimate layer of this classifier as the feature space for anomaly detection.

\item Our novel \texttt{MCIF} method addresses the limitations of using a single isolation forest on the complex latent space by training separate isolation forests for each known transient class and taking the minimum score as the final anomaly score.

\end{enumerate}

A significant contribution of this work is the demonstration that a well-trained classifier can be effectively repurposed for anomaly detection by leveraging the clustering properties of its latent space. The flexibility of our approach allows for the adaptation of any classifier to an anomaly detector. For example, using existing classifiers as feature extractors for astronomical spectra, images, or time series from other domains, we can build effective anomaly detectors.

\bibliography{example_paper}
\bibliographystyle{icml2024}

%%%%%%%%%%%%%%%%%%%%%%%%%%%%%%%%%%%%%%%%%%%%%%%%%%%%%%%%%%%%%%%%%%%%%%%%%%%%%%%
%%%%%%%%%%%%%%%%%%%%%%%%%%%%%%%%%%%%%%%%%%%%%%%%%%%%%%%%%%%%%%%%%%%%%%%%%%%%%%%
% APPENDIX
%%%%%%%%%%%%%%%%%%%%%%%%%%%%%%%%%%%%%%%%%%%%%%%%%%%%%%%%%%%%%%%%%%%%%%%%%%%%%%%
%%%%%%%%%%%%%%%%%%%%%%%%%%%%%%%%%%%%%%%%%%%%%%%%%%%%%%%%%%%%%%%%%%%%%%%%%%%%%%%
\newpage
\appendix
\onecolumn

\section{Visual Comparison to other Approaches}

\begin{figure*}
    \centering

    \includegraphics[width=\textwidth]{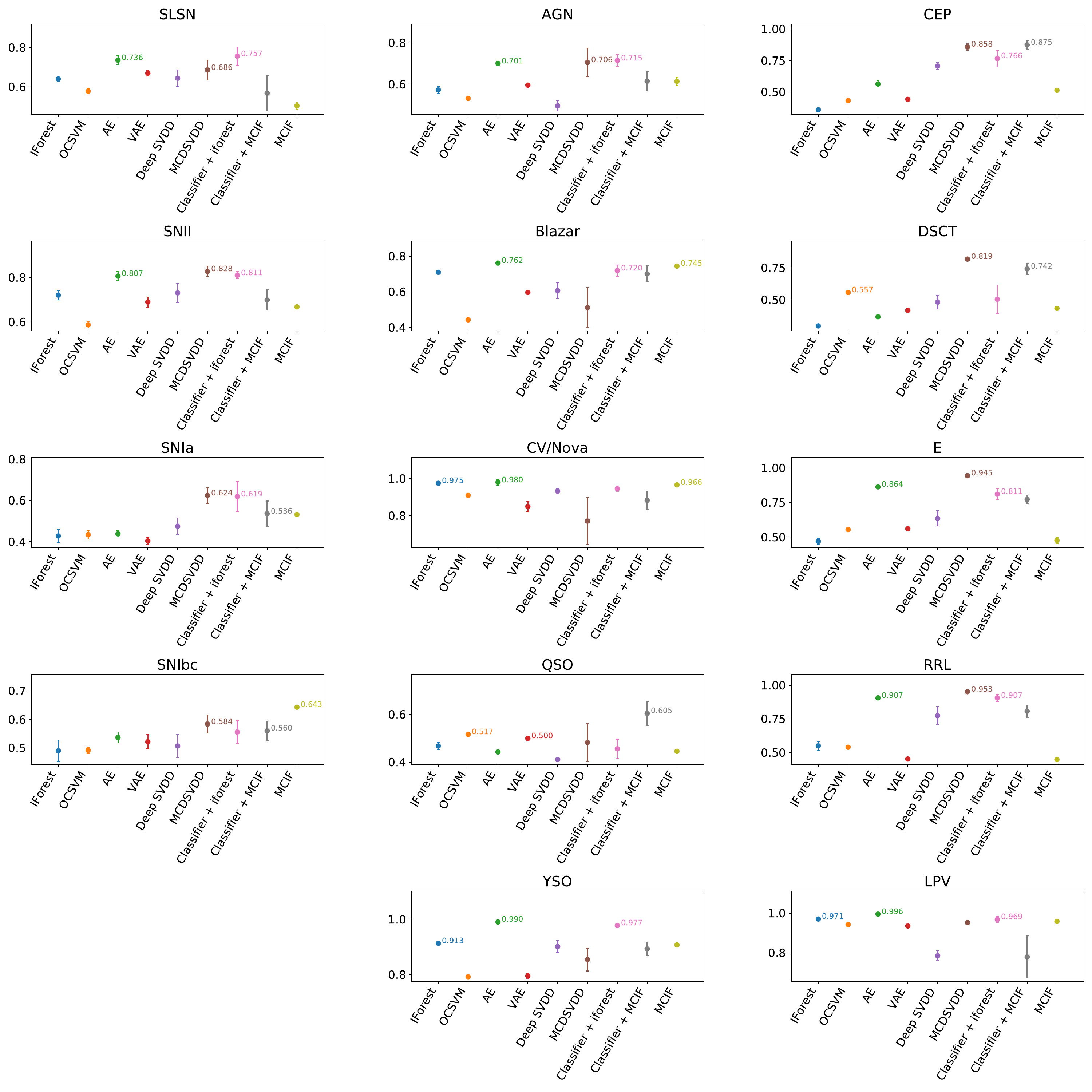}

    \caption{Visual representation of the comparative analysis depicted in Table \ref{table:results}. The AUROC is written for the models top 3 models for each class.}
    \label{fig:visual_table}
\end{figure*}

Figure \ref{fig:visual_table} is a visual representation of the results depicted in Table \ref{table:results}.

\section{Dataset Information}
\label{sec:dataset}

\begin{figure*}
    \centering
    \begin{tabular}{ll}
      \includegraphics[scale=0.28]{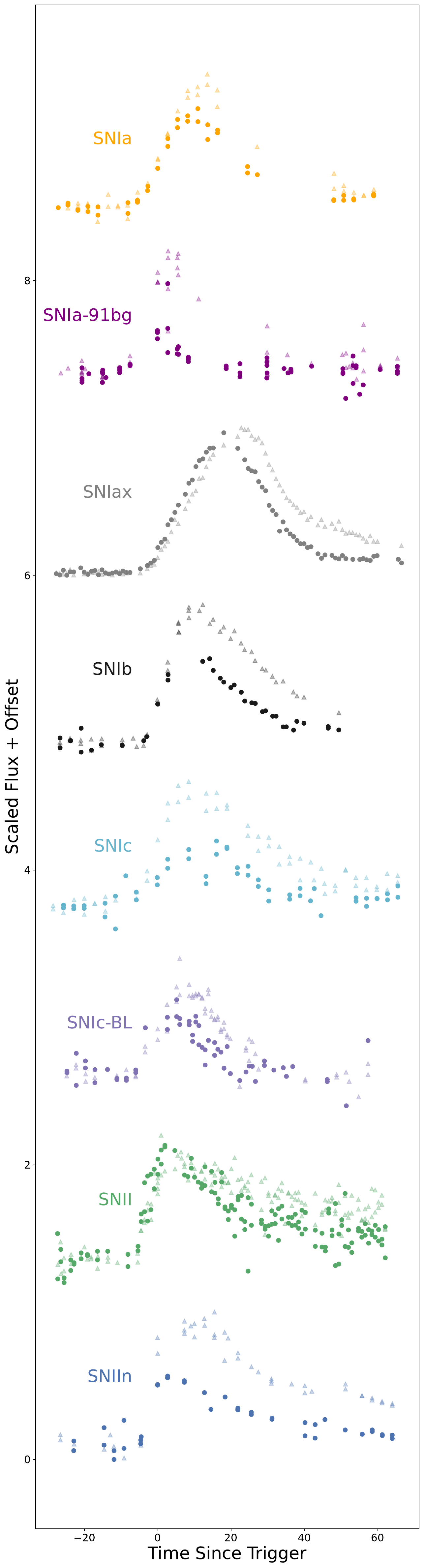}  
      &
      \includegraphics[scale=0.28]{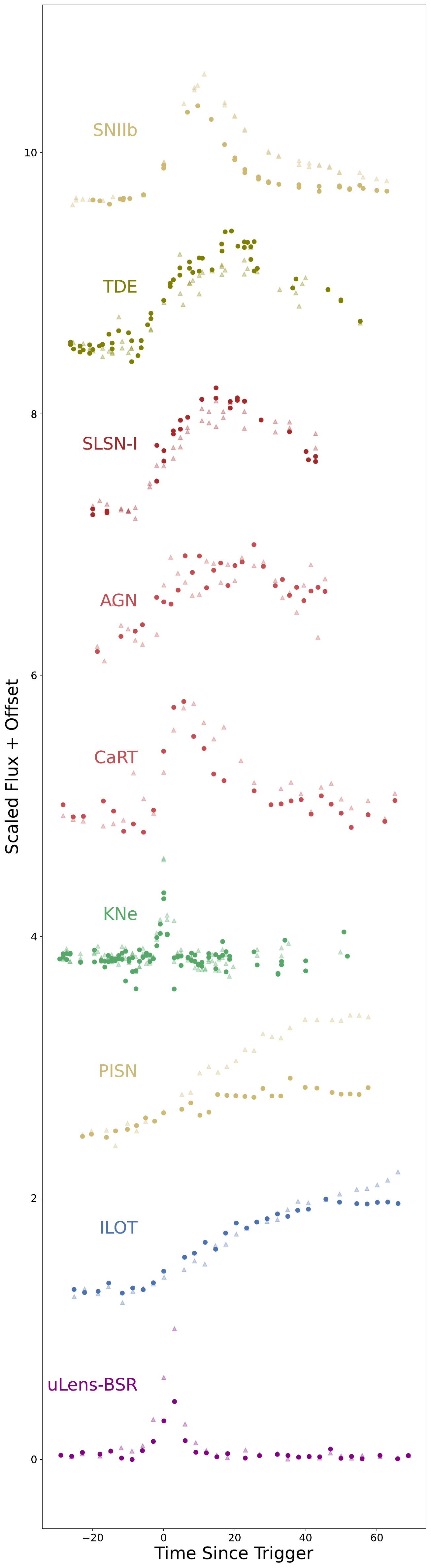}  
    \end{tabular}
    
    \caption{Sample light curves from each transient class used in this work. We only plot transients with low signal-to-noise to help visually compare shapes. The dark circular markers represent the r band while the light triangular markers represent the g band. Flux errors are not plotted.}
    \label{fig:samplecurves}
\end{figure*}

A sample light curve from each class is illustrated in Figure \ref{fig:samplecurves}. Table \ref{table:populationdistribution} reports the number of objects from each class in our training set and realistic sample used for evaluation in Section \ref{sec:representative_population}.

% The distribution of the mean cadence for each class is shown in Figure \ref{fig:cadence} and we verify that there are no irregularities that could be assisting our model.

\begin{table}[]
\begin{center}

\caption{Number of transients in the training set, validation set, test set, and realistic samples (see section \ref{sec:representative_population}) for each class. All anomalous data is reserved for evaluation.}
\label{table:populationdistribution} 
\begin{tabular}{||c c c c c c||} 
 \hline
 Class & Training & Validation & Test & Total & \specialcell{Realistic\\Sample$^a$} \\ [0.5ex] 
 \hline\hline
SNIa & 9314 & 1131 & 1142 & 11587 & 1142 \\
\hline
SNIa-91bg & 10361 & 1318 & 1321 & 13000 & 1318 \\
\hline
SNIax & 10413 & 1248 & 1339 & 13000 & 1339 \\
\hline
SNIb & 4197 & 507 & 563 & 5267 & 563 \\
\hline
SNIc & 1279 & 169 & 135 & 1583 & 135 \\
\hline
SNIc-BL & 1157 & 124 & 142 & 1423 & 142 \\
\hline
SNII & 10420 & 1279 & 1301 & 13000 & 1301 \\
\hline
SNIIn & 10323 & 1359 & 1318 & 13000 & 1318 \\
\hline
SNIIb & 9882 & 1233 & 1208 & 12323 & 1208 \\
\hline
TDE & 9078 & 1162 & 1114 & 11354 & 1114 \\
\hline
SLSN-I & 10285 & 1322 & 1273 & 12880 & 1273 \\
\hline
AGN & 8473 & 1046 & 1042 & 10561 & 1042 \\
\hline
CaRT & 0 & 0 & 10353 & 10353 & $11 \pm 3$ \\
\hline
KNe & 0 & 0 & 11166 & 11166 & $11 \pm 3$ \\
\hline
PISN & 0 & 0 & 10840 & 10840 & $11 \pm 3$ \\
\hline
ILOT & 0 & 0 & 11128 & 11128 & $10 \pm 3$ \\
\hline
uLens-BSR & 0 & 0 & 11244 & 11244 & $10 \pm 3$ \\
\hline

\multicolumn{6}{l}{$^a$ The mean number of transients across the 50 test samples is shown. The}\\
\multicolumn{6}{l}{errors refer to the STD in the population size across the 50 sets. All common}\\
\multicolumn{6}{l}{ test data is part of every sample, hence errors are not shown.}\\
\end{tabular}

\end{center}
\end{table}

% \begin{figure*}
%     \centering
%   \includegraphics[scale=0.5]{images/cadencedist.pdf}  

%     \caption{Distribution of mean cadence for all classes. We verify that there are no irregularities that could be assisting our model.}
%     \label{fig:samplecurves}
% \end{figure*}
% \section{Using a Preexisting Classifier}

% We show that a classifier with a small penultimate layer (9 neurons in the text) can be effectively used for anomaly detection, however preexisting classifiers with larger penultimate layers can also be used.  

\section{Classifier Results}
\label{sec:classifier}
\begin{figure*}
\centering
\begin{tabular}{ll}
\includegraphics[width=0.5\textwidth]{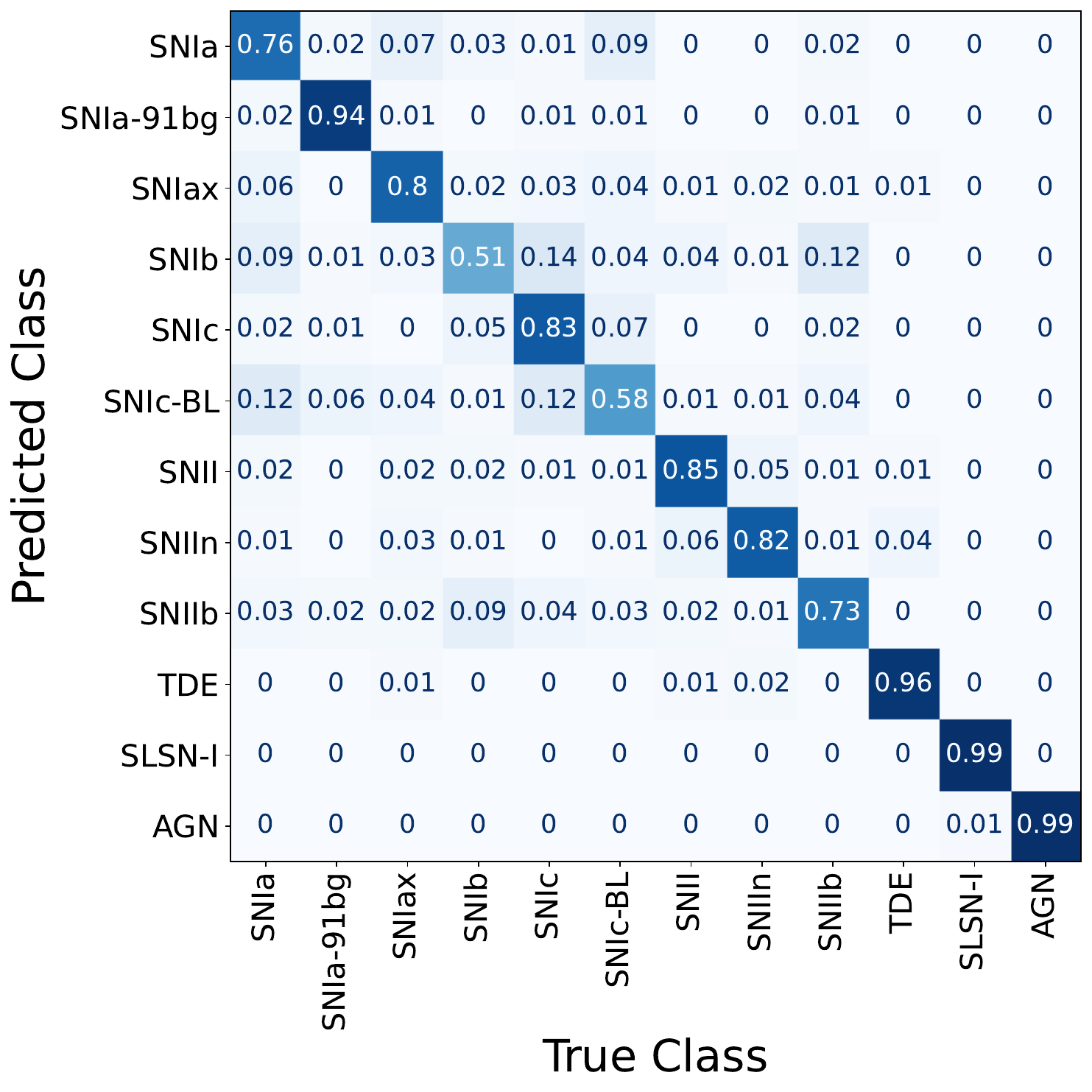}
&
\includegraphics[width=0.5\textwidth]{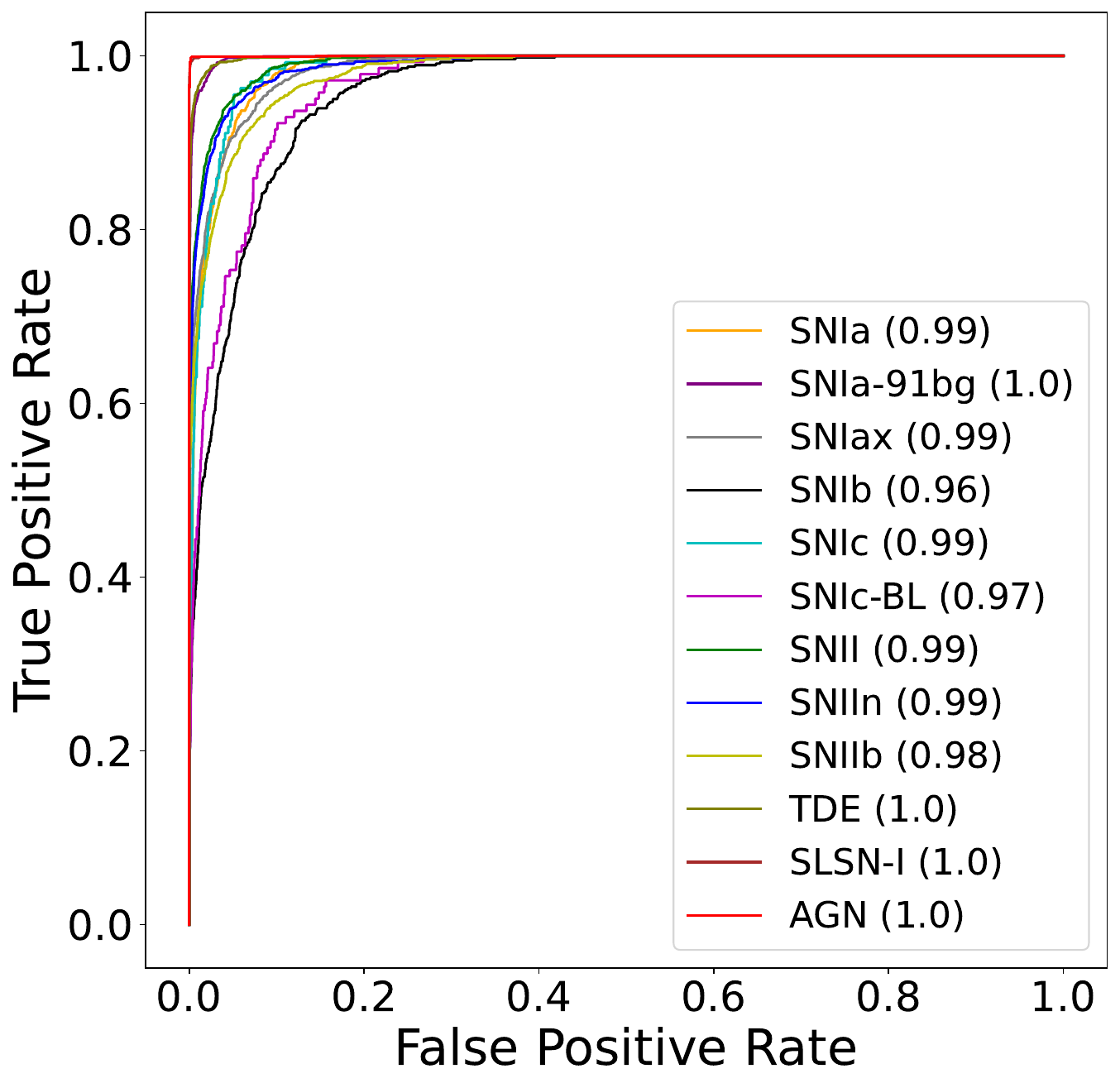}
\end{tabular}
\caption{The normalized confusion matrix [left] and ROC curve [right] of the 12 common transient classes used for training given full light curve data. Each cell in the confusion matrix signifies the fraction of transients from each \textit{True Class} that was classified into the \textit{Predicted Class}. The ROC curve illustrates the True Positive Rate against the False Positive Rate across various threshold probabilities for each class, with the Area Under ROC curve (AUROC) in parenthesis. The model's evaluation is conducted on the test set consisting of 10\% of the data from the common classes.}
\label{fig:ConfusionROC}
\end{figure*}

The normalized confusion matrix in Figure \ref{fig:ConfusionROC} [left] illustrates our classifier's ability to accurately predict the correct transient class on the test data. Each cell indicates the fraction of transients from the true class that are classified into the predicted class. The high values along the diagonal, approaching 1.0, indicate strong performance.  The misclassifications, indicated by the off-diagonal values, predominantly occur between subclasses of Type Ia supernovae (SNIa, SNIa-91bg and SNIax) and between the core-collapse supernova types (SNIb, SNIc, SNII subtypes), which is expected given their observational similarities. These SNe have been shown to confuse previous models (see Fig. 7 of \citealp{Muthukrishna19RAPID}). 

\section{Real-Time Detection}

Identifying anomalies in real-time is important for obtaining early-time follow-up observations, which is crucial for understanding their physical mechanisms and progenitor systems \citep[e.g.][]{Kasen2010}. However, directly assessing our architecture's real-time performance is challenging due to the irregular sampling of light curves in our input format.

\begin{figure*}
  \centering

  \begin{tabular}{ll}
  \includegraphics[scale=0.40]{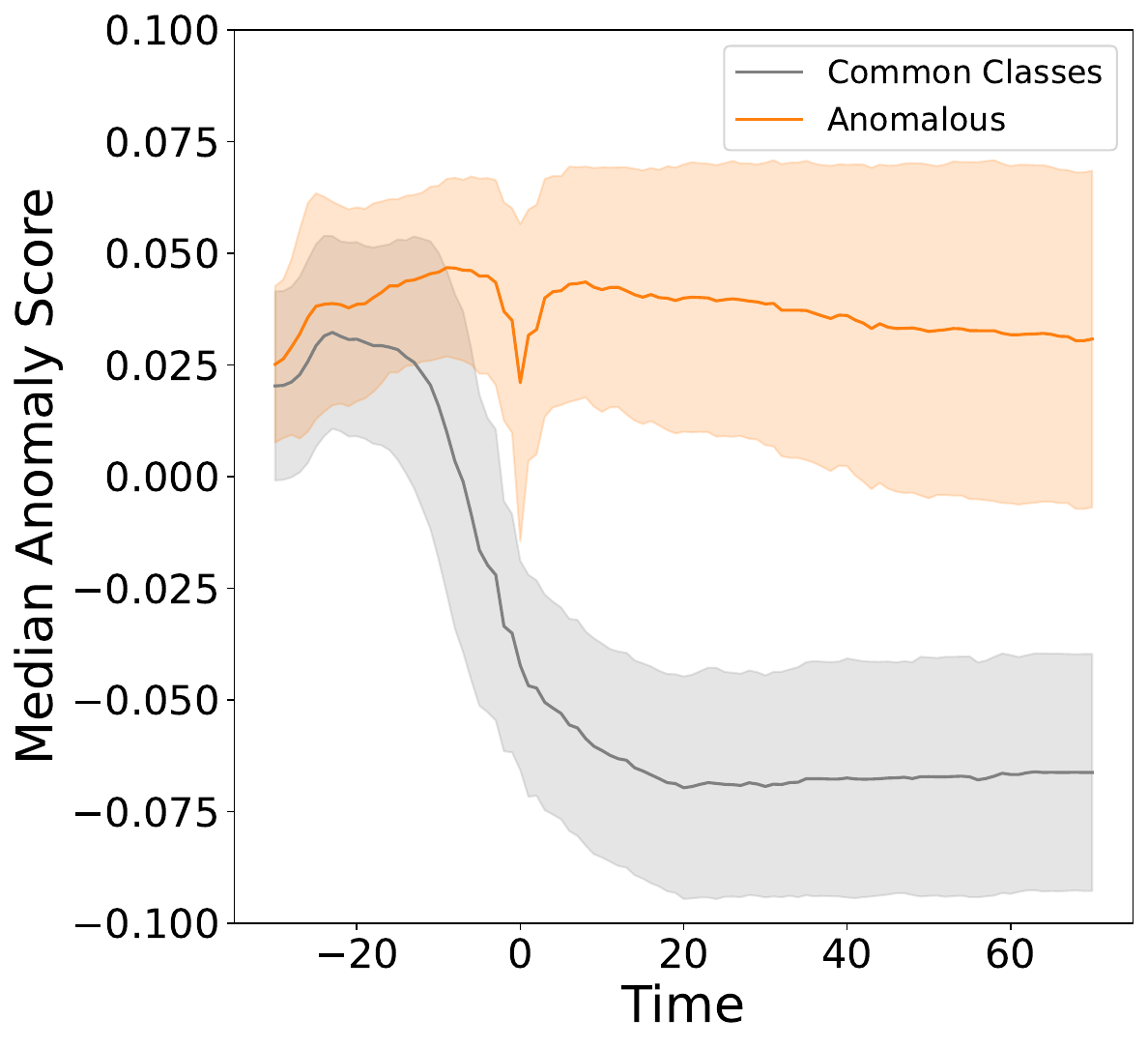}  
  &
  \includegraphics[scale=0.40]{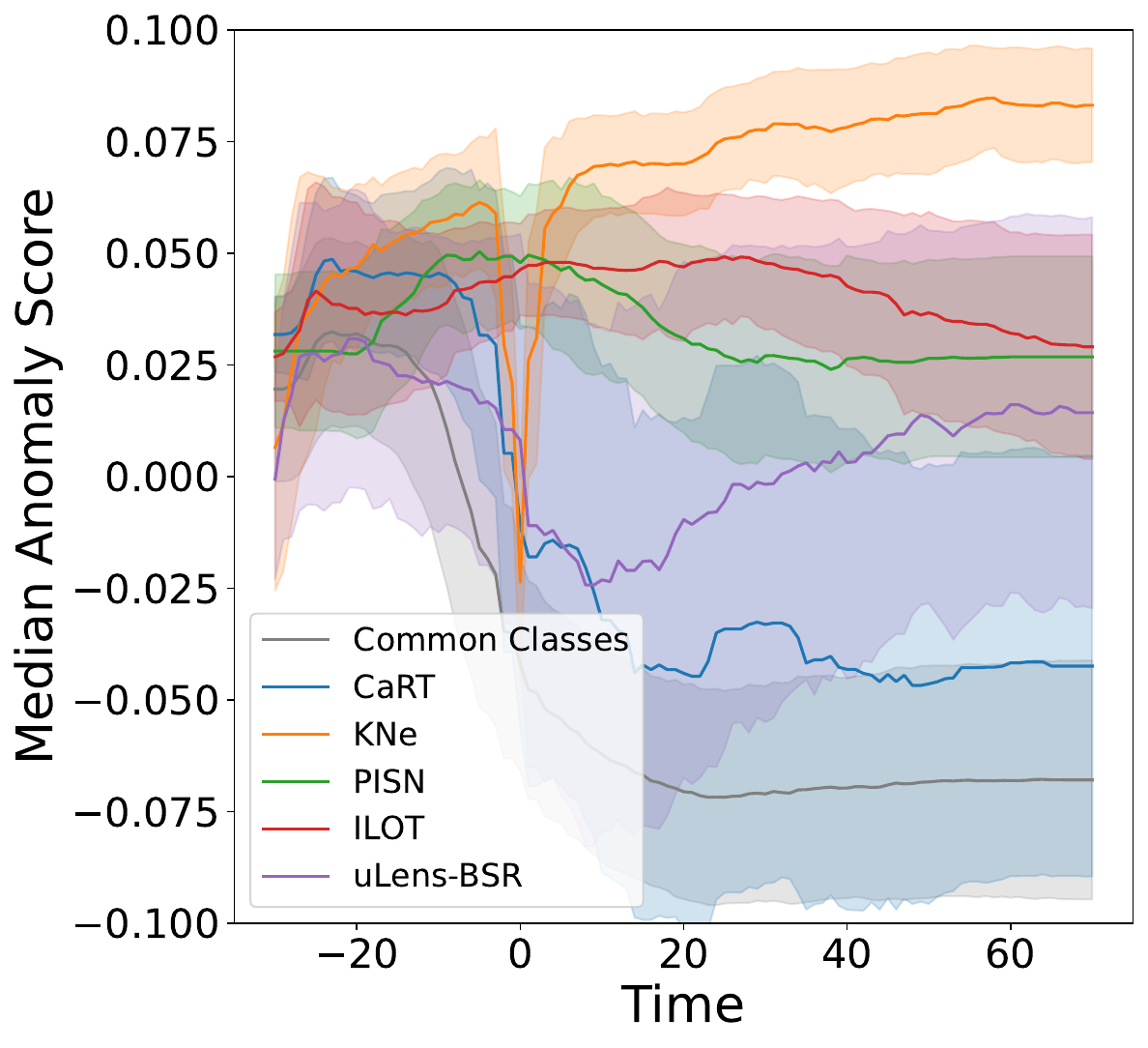}  
  \end{tabular}

  \caption{Median \texttt{MCIF} anomaly score over time for a sample of transients from the test set. Real-time anomaly scores are calculated at intervals of 1 day for a sample of 2000 common and 2000 total anomalous light curves. The left plot shows the scores for the common and anomalous transients as a whole, while the right plot shows each anomalous class individually. The anomaly scores for the common transients decline before the trigger, while the anomalous transients remain at high scores throughout most of the transient's evolution.}
  \label{fig:RealTimeAnalysis}
\end{figure*}

To assess the real-time performance of our architecture, we plot the median anomaly scores over time for a sample of 2000 common and 2000 anomalous transients in Figure \ref{fig:RealTimeAnalysis}. To construct this plot without relying on interpolation, we calculate scores at discrete times $l$ sampled at 1-day intervals from $-30$ to $70$ days relative to trigger, using only observations occurring before each time $l$ to mimic a real-time scenario. To ensure sufficient information for robust scoring, we only consider transients where the final observation was recorded after time $l - 5$. The results show a clear divergence where common transient scores tend to decline around trigger, while anomalous transient scores remain consistently high. 

Figure \ref{fig:RealTimeAnalysis} reveals two notable irregularities. Firstly, the anomaly scores for common transients decline before trigger, which is unexpected given that the pre-trigger phase of most transient classes should primarily consist of background noise. Further analysis of the pre-trigger classification results reveals that certain transients, most notably SLSN-I and AGN, are almost all classified before trigger, thereby lowering the average anomaly score for common transients. This can be attributed to the fact that redshift and pre-trigger information such as host galaxy color and some AGN pre-trigger variability are particularly useful for classifying these transients before trigger (see Figure 16 of \citealp{Muthukrishna19RAPID}).

Secondly, KNe exhibit a significant dip around the time of trigger. Upon further analysis, we found that certain common transient classes also experienced a similar dip around trigger; however, unlike KNe, they do not rebound back to higher anomaly scores. This dip is related to the inherent nature of the trigger of a light curve, which often marks the first \textit{real} observation of the transient phase of a light curve, and serves as a reset for the anomaly score. A more detailed analysis of this phenomenon is omitted for brevity.

These preliminary findings suggest the potential for enabling real-time identification of anomalous transients. While some known rare classes can be difficult to distinguish from the common classes without a significant amount of data, others can be detected surprisingly soon after trigger. The ability to flag unusual events early in their evolution could prove invaluable for optimizing the allocation of follow-up resources and maximizing the scientific returns from rare transient discoveries.

\end{document}